\documentclass{article}

\usepackage[preprint]{neurips_2024}

\usepackage[utf8]{inputenc} 
\usepackage[T1]{fontenc}    
\usepackage{url}            
\usepackage{booktabs}       
\usepackage{amsfonts}       
\usepackage{nicefrac}       
\usepackage{microtype}      
\usepackage{xcolor}         
\usepackage{tabularx} 
\usepackage{booktabs}
\usepackage{pifont} 
\usepackage{multirow} 
\usepackage{array} 
\usepackage{amssymb}
\usepackage{xcolor}
\usepackage{graphicx}
\usepackage{appendix}
\usepackage{enumitem}
\usepackage{algorithm}
\usepackage{algpseudocode}
\usepackage{ifthen}
\usepackage{amsmath} 
\usepackage[colorlinks=true, linkcolor=blue, citecolor=blue, urlcolor=blue]{hyperref}
\usepackage{cleveref}

\usepackage{natbib}

\crefname{section}{\S}{\S}
\Crefname{section}{\S}{\S}
\crefformat{section}{(§#2#1#3)}
\Crefformat{section}{(§#2#1#3)}

\definecolor{checkmarkcolor}{HTML}{008000}   
\definecolor{crossmarkcolor}{HTML}{FF0000}  
\newcommand{\cmark}{\textcolor{checkmarkcolor}{\ding{51}}}   
\newcommand{\xmark}{\textcolor{crossmarkcolor}{\ding{55}}}

\title{ROMAS: A \underline{Ro}le-Based \underline{M}ulti-\underline{A}gent \underline{S}ystem for Database monitoring and Planning}

\author{%
Yi Huang$^{1}$\thanks{Equal contribution}, Fangyin Cheng$^{2}$\footnotemark[1], Fan Zhou$^{1}$, Jiahui Li$^{1}$,\\Jian Gong$^{1}$, Hongjun Yang$^{1}$, Zhidong Fan$^{1}$, Caigao Jiang$^{1}$,\\Siqiao Xue$^{1}$\thanks{Corresponding Author}, Faqiang Chen$^{1}$\footnotemark[2]\\$^{1}$Ant Group, $^{2}$JD Group \\
\texttt{\{yaqing.hy, faqiang.cfq\}@antgroup.com}, \texttt{siqiao.xsq@gmail.com}
}

\begin{document}

\maketitle

\begin{abstract}

In recent years, Large Language Models (LLMs) have demonstrated remarkable capabilities in data analytics when integrated with Multi-Agent Systems (MAS). However, these systems often struggle with complex tasks that involve diverse functional requirements and intricate data processing challenges, necessitating customized solutions that lack broad applicability. Furthermore, current MAS fail to emulate essential human-like traits such as self-planning, self-monitoring, and collaborative work in dynamic environments, leading to inefficiencies and resource wastage. To address these limitations, we propose ROMAS, a novel \underline{Ro}le-Based \underline{M}ulti-\underline{A}gent \underline{S}ystem designed to adapt to various scenarios while enabling low code development and one-click deployment. ROMAS has been effectively deployed in DB-GPT~\citep{xue2023dbgpt,xue2024demonstration}, a well-known project utilizing LLM-powered database analytics, showcasing its practical utility in real-world scenarios. By integrating role-based collaborative mechanisms for self-monitoring and self-planning, and leveraging existing MAS capabilities to enhance database interactions, ROMAS offers a more effective and versatile solution. Experimental evaluations of ROMAS demonstrate its superiority across multiple scenarios, highlighting its potential to advance the field of multi-agent data analytics.
\end{abstract}

\section{Introduction}

\begin{table}[ht] 
\centering 
\setlength{\tabcolsep}{2pt}   
\small   
\begin{tabularx}{0.8\textwidth}{lccccccc} 
\toprule 
Perspective & Module & Function & ROMAS & GA & AA & MG  \\  
\midrule 
\multirow{9}{*}{Structure  design}  & \multirow{2}{*}{Profile}  & Role-based  cooperation  & \cmark  & \cmark  & \cmark  & \cmark  \\
& & Scenario  versatility  & \cmark  & \cmark  & \cmark  & \cmark  \\
\cmidrule(lr){2-8}  \cmidrule(lr){2-7} 
&\multirow{3}{*}{Task  planning}  & Global monitor  & \cmark  & \xmark  & \cmark  & \xmark  \\
&  & Dynamic agent generation & \cmark  & \cmark  & \cmark  & \xmark  \\
& & Plan corrected  & \cmark  & \xmark  & \xmark  & \xmark  \\
\cmidrule(lr){2-8}  \cmidrule(lr){2-7} 
&\multirow{2}{*}{Memory}  & Message  queue  & \cmark  & \cmark  & \xmark  & \xmark  \\
& & Hybrid  memory  & \cmark  & \cmark  & \cmark  & \cmark  \\
\cmidrule(lr){2-8}  \cmidrule(lr){2-7} 
&\multirow{2}{*}{Action}  & Guardrails  & \cmark  & \cmark  & \cmark  & \cmark  \\
& & Self-reflection  & \cmark  & \cmark  & \cmark  & \cmark  \\
\cmidrule(lr){1-8}  \cmidrule(lr){2-7} 
\multirow{3}{*}{Scenario adaptation} & \multirow{3}{*}{-}  & Versatility & \cmark  & \cmark  & \cmark  & \xmark  \\
&  & Flexibility & \cmark  & \cmark  & \cmark  & \cmark  \\
&  & Robustness & \cmark  & \xmark  & \xmark  & \xmark  \\
\bottomrule 
\end{tabularx} 
\vspace{0.3cm}  
\caption{Comparison of ROMAS and traditional Autonomous MAS (GA = Generative Agent, AA = AutoAgents, MG = MetaGPT).}
\label{tab:mas_comp} 
\end{table}
Multi-agent systems (MAS) have garnered significant attention for their potential to tackle complex tasks in dynamic environments through the coordinated collaboration of multiple agents. As task complexity and environmental dynamics increase, researchers have been working to enhance the capabilities of MAS by decomposing intricate tasks into simpler subtasks and improving agents' adaptive, reflective, and self-correcting abilities ~\citep{anderson2018evaluation,Wang2023VoyagerAO,xue2023dbgpt,trivedi2024appworld}. However, despite these efforts, current MAS approaches still face several critical limitations.

In terms of structural design, traditional MAS often rely on static task allocation and predefined processes, such as Chain of Thought (CoT)~\citep{wei2022chain}, Self-consistent CoT (CoT-SC)~\citep{wang2023selfconsistencyimproveschainthought} and Tree of Thought (ToT)~\citep{yao2023tree}. These procedural methods suffer from low fault tolerance and lack the capability for autonomous reflection and interaction, leading to failures when deviations occur from the predetermined plan. Furthermore, task-oriented MAS, exemplified by frameworks such as MetaGPT~\citep{hong2023metagpt} and TaskWeaver~\citep{qiao2024taskweavercodefirstagentframework}, are domain-specific and fail to generalize well beyond their predefined scopes, thus limiting their applicability. Interactive MAS, such as ReAct~\citep{yao2023reactsynergizingreasoningacting} , ChatCoT~\citep{chen2023chatcottoolaugmentedchainofthoughtreasoning} and Voyager~\citep{Wang2023VoyagerAO}, facilitate dynamic feedback and self-correction but struggle with high correction costs and the inability to rectify previously executed subtasks.

In terms of implementation, traditional MAS application development often relies on open-source frameworks, such as LangChain~\citep{langchain}, Rasa\cite{rasa2023}, ChatDev~\citep{qian2024chatdevcommunicativeagentssoftware}, and AgentScope~\citep{gao2024agentscopeflexiblerobustmultiagent}, which offer a range of development tools, simulated environments, and fundamental agent capabilities. Although these frameworks allow for the quick development of customized MAS systems, they are limited by their inability to handle diverse data management, provide robust built-in development components, and offer comprehensive complex problem solving toolkits, which hinders the full development potential of MAS applications. 

To address the aforementioned limitations, we introduce ROMAS, a novel role-based multi-agent system deployed in DB-GPT\footnote{\url{https://github.com/eosphoros-ai/DB-GPT}}, featuring several key innovations.

\begin{itemize}[leftmargin=*]
    \item \textbf{Role-based collaboration}. ROMAS organizes agents into specific roles—planner, monitor, and worker. The planner creates global task lists and allocates subtasks to workers~\cref{sec:init}. The monitor oversees workers, ensuring correctness and re-planning when needed~\cref{sec:execute}. This structure enables real-time self-supervision, enhancing flexibility and robustness.
    \item \textbf{Self-monitoring and self-planning}. ROMAS allows agents to evaluate their performance and adjust actions dynamically through self-monitoring and self-planning mechanisms, ensuring high adaptability to changing conditions and complex tasks.
    \item \textbf{Low-code development and one-click deployment}. We developed ROMAS on the DB-GPT, a multi-agent application framework that integrates efficient computing operators, rich database management components, user-friendly data analysis visualization, flexible multi-domain deployment, and the Agentic Workflow Expression Language (AWEL) for low-code streamlined development. DB-GPT provides users with an efficient, intuitive, and secure data interaction solution, facilitating development and simplifying deployment.   
    \item \textbf{Enhanced database interactions}. ROMAS optimizes data retrieval, processing, and storage, with the help of DB-GPT's advanced data handling capabilities. This makes ROMAS ideal for applications involving large datasets and complex analytics, ensuring efficient and effective data management.
\end{itemize}

\section{Related Work}

\begin{table}[ht] 
\centering 
\setlength{\tabcolsep}{2pt} 
\small   
\begin{tabularx}{0.8\textwidth}{lccccccc} 
\toprule 
Module  & Component  & DB-GPT  & AgentScope  & AutoGen & LangChain  \\
\midrule 
 \multirow{4}{*}{Rich component}  & Multiple  databases  & \cmark  & \cmark  & \xmark  & \xmark  \\
& Data analysis  & \cmark  & \cmark  & \cmark  & \cmark  \\
& LLM proxy  & \cmark  & \cmark  & \cmark  & \cmark  \\
& GraphRAG & \cmark  & \cmark  & \xmark  & \xmark  \\
\cmidrule(lr){2-8}  \cmidrule(lr){1-6} 
\multirow{3}{*}{Fine-tuning} & Text2SQL  & \cmark  & \xmark  & \xmark  & \xmark  \\ 
& NLU  & \cmark  & \xmark  & \xmark  & \xmark  \\
& Prompt  & \cmark  & \cmark  & \cmark  & \cmark \\
\cmidrule(lr){2-8}  \cmidrule(lr){1-6}
\multirow{4}{*}{Coding}  & Workflow language& \cmark  & \cmark  & \cmark  & \cmark  \\
& Operator & \cmark  & \xmark  & \cmark  & \cmark  \\
& Private deployment & \cmark  & \xmark  & \cmark  & \xmark  \\
& Distribution & \cmark  & \cmark  & \cmark  & \xmark \\
\cmidrule(lr){2-8}  \cmidrule(lr){1-6} 
\multirow{2}{*}{User interaction} & Drag-and-drop workshop  & \cmark  & \cmark  & \xmark  & \xmark  \\
& Visualization page  & \cmark  & \cmark  & \cmark  & \xmark  \\
\bottomrule 
\end{tabularx} 
\vspace{0.3cm}  
\caption{Comparison  of  DB-GPT and  traditional agent application framework.}
\label{tab:app_comp} 
\end{table}

Recent research in MAS has focused on three key areas: structure design, application development frameworks, and scenario adaptation~\citep{Wang_2024,jiang2023anytime,jiang2024interpretable}. Structure design emphasizes enhancing agent performance through modules such as profile, memory, planning, and action~\citep{masterman2024landscapeemergingaiagent}. Application development framework emphasis on achieving low-code development, convenient deployment, and user interaction experience. Scenario adaptability emphasis on the ability to be effectively applied across various domains. A highly adaptable framework should be able to easily accommodate different task requirements without requiring extensive adjustments. 

\textbf{Autonomous MAS.} We compare ROMAS with traditional MAS in terms of structural design and scenario adaptation as shown in table \ref{tab:mas_comp}, Generative  Agents~\citep{park2023generativeagentsinteractivesimulacra} simulates  social  role  definitions,  determining  each  agent's  behavior,  tasks,  and  interaction  modes.  The  memory  module  records  all  experiences  through  a memory  stream  and  retrieves  the  highest-priority  memories  based  on  recency,  importance,  and  relevance. However,  in  task  planning,  Generative  Agents  lack  a global  monitor  mechanism,  limiting  ability  to  engage  in  global  reflection and dynamic correction. AutoAgents~\citep{chen2024autoagentsframeworkautomaticagent} adaptively generate specialized agents to build an AI team, which consists of two critical stages: drafting stage and execution stage.  During the drafting stage, the planner determines the plan list and the generation of agents through discussions with the agent observer and the plan observer. In the execution stage, the action observer corrects the behavior of individual agents. Although multiple global supervisory roles are set, there is no correction of the plan list during the execution stage. Therefore, if there is an error in the plan list during the drafting stage, it cannot be corrected in the execution stage. MetaGPT~\citep{hong2023metagpt} assigns  multiple  engineer  agent roles to  collaboratively  complete  the  software  development  coding  and  Standard  Operating  Procedure  (SOP) writing. However,  MetaGPT  lacks a global monitor and  can't adaptively generate specialized agents.  This means that once the plan is set, MetaGPT has limited flexibility for adjustments or corrections during execution. 

\textbf{Application framework.} We compare DB-GPT with traditional agent application framework as shown in table \ref{tab:app_comp}, AgentScope~\citep{gao2024agentscopeflexiblerobustmultiagent} is a developer-centric multi-agent platform that offers user-friendly interfaces for application demonstration and monitoring, a zero-code programming workstation, and an automatic prompt tuning mechanism. In terms of coding, it provides rich syntactic tools, built-in and customizable fault tolerance mechanisms, and an actor-based distributed framework. However, AgentScope lacks components such as Text2SQL fine-tuning~\citep{zhou2024dbgpthub} for small models and  ready-made operators, making it less suitable for data analysis compared to DB-GPT. LangChain~\citep{langchain}  offers a rich set of components, such as LLMs, memory, and agents, as well as structured component collections like chains to accomplish specific tasks. However, LangChain lacks a user-friendly interface and does not support private deployment. AutoGen~\citep{wu2023autogenenablingnextgenllm} main tasks include: (1) defining conversational agents with specific capabilities and roles; and (2) programming the interaction behaviors between agents through computation and control within a dialogue center. However, programming based on a QA scenario has limitations when it comes to handling complex tasks.

\section{Methodology}
\label{sec:methodology}
\begin{figure}[t] 
\centering
\includegraphics[width=0.7\linewidth]{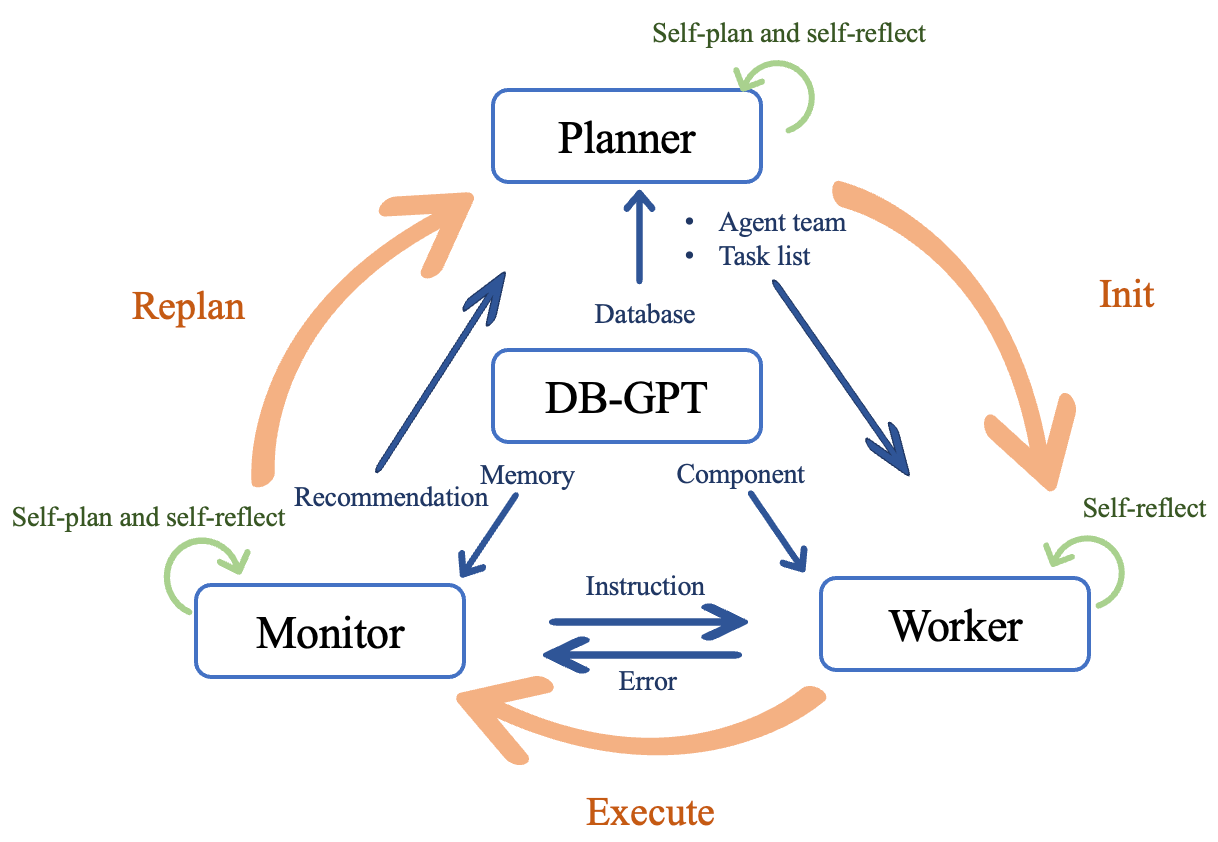} 
\caption{ROMAS framework. The blue lines represent key message exchanged between agents and DB-GPT, the orange lines signify the three distinct phases of ROMAS, and the green lines denote each agent's internal planning and reflection processes.} 
\label{fig:romas_framework} 
\end{figure}
ROMAS is a versatil data analysis framework based on the DB-GPT. Its principles prioritize high flexibility and comprehensive functionality, without being constrained by specific scenarios, data formats, or development complexity. the agent roles in ROMAS are divided into a planner, a monitor, and multiple workers~\citep{5345731} shown in figure\ref{fig:romas_framework}, a detailed description of the role definition is provided in the appendix\ref{role_intro}.

The ROMAS comprises three critical phases: \textbf{initialization}, \textbf{execution}, and \textbf{re-planning}, each benefiting from the powerful database capabilities of DB-GPT. During the initialization phase, the planner decomposes the requirements based on the scenario description and known database information, forming an specailized agent team and designing specific task lists for each agent~\citep{li2024agentorientedplanningmultiagentsystems}. The execution phase relies on multi-agent cooperation~\citep{du2023reviewcooperationmultiagentlearning}, with workers collaborating to complete tasks according to the planner's plan. If errors occur, the system reports key global information to the monitor, awaiting further instructions. In the re-planning phase, the monitor and planner interact closely. If the monitor's attempts to correct the errors fail, it triggers the planner to re-plan, integrating key information to assist in the planner's decision-making.

\subsection{Initialization Phase}\label{sec:init}
\begin{figure}[ht] 
\centering
\includegraphics[width=\linewidth]{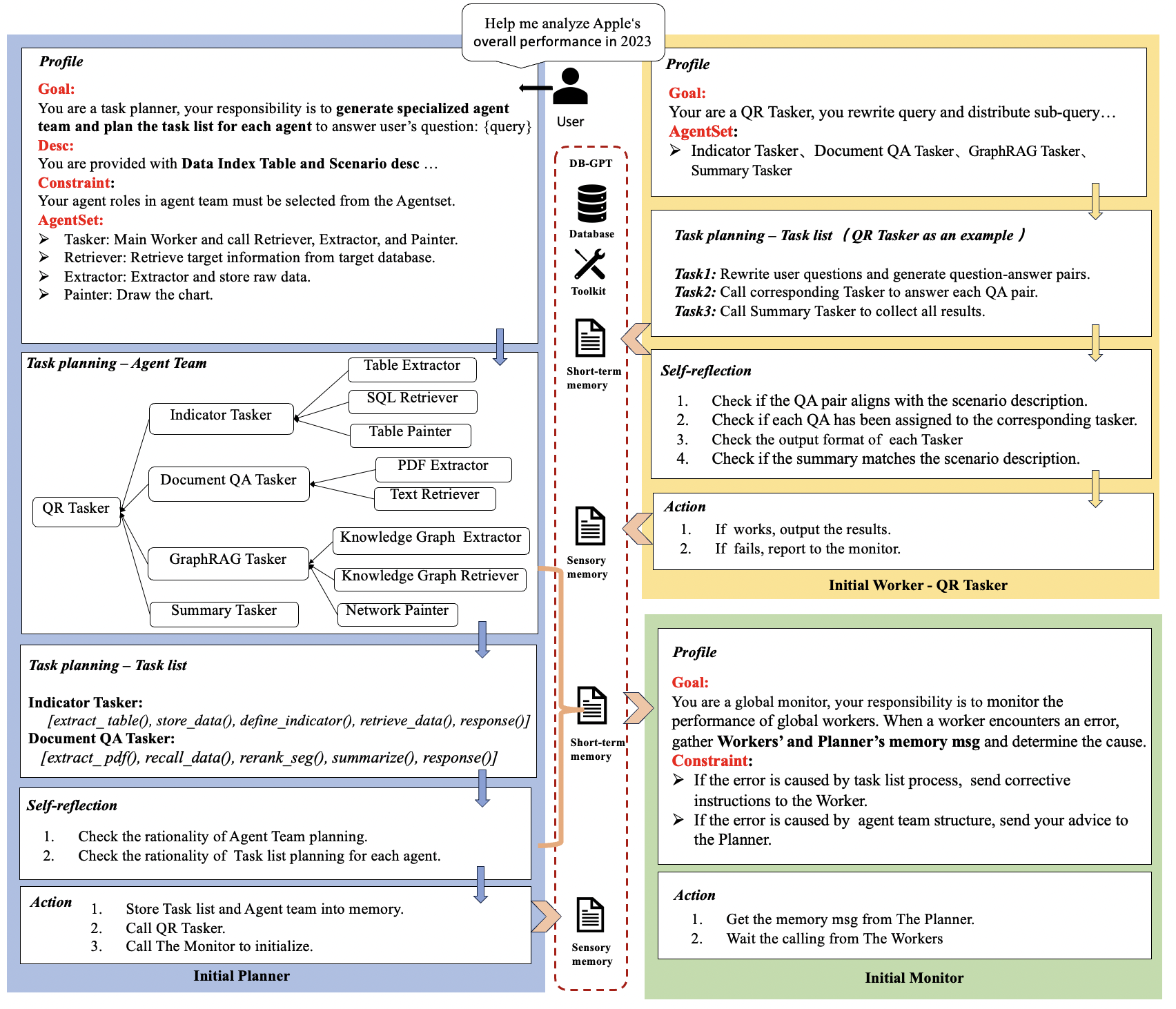} 
\caption{Initialization phase of ROMAS, the planner is primarily responsible for two steps: self-planning and self-reflection. In self-planning, the planner generates the agent team and arranges subtask list. In self-reflection, these strategies are validated for logical consistency.} 
\label{fig:initial_stage} 
\end{figure}
In this phase as illustrated in figure \ref{fig:initial_stage}, firstly the planner automatically generate the profile prompt based on templates and user input. Next, the planner will execute the self-planning process according to the profile. During this process, it will sequentially generate two key strategies: the cooperation workflow of the agent team and the task list for each agent. Once the strategies are generated, the planner will perform self-reflection process to check the effectiveness of these strategies. Finally, an action is executed to save the initial strategies and trigger the execution phase. 

Self-planning process~\citep{huang2024understandingplanningllmagents}. The input set {<G, D, C, A, T>} is defined in the profile. and the output set is {<AL, TL>} which is the input of self-reflection process. Goal G specifies the planner’s task goal in detail, which involve creating the entire cooperative agent team and assigning refined task lists for each worker. Description D describes the provided scenario information, including descriptions from user and a brief table with raw database index information for task assigning. Constraint C represents the constraints that the planner must adhere to when generating the agent team and task lists. Toolkit T is a set of pre-defined toolkits provided by DB-GPT like web search tool~\citep{bing}. Multiple tools can be combined to complete specific tasks in the agent's task list. Agentset A defines the range of agent roles available for generating the agent team. worker roles are limited to tasker, retriever, extractor, and painter. Each role has a pre-defined parent class template that includes many general capabilities, which can be directly inherited and further refined based on the specific scenario. 

Agent list AL is a list of agents with a predefined call order, structured like a tree~\citep{yao2023tree}. tasker serves as the control center for each entire process including main process, planning and managing the workflow of each agent. Following steps are executed in sequence in the tasker: 1. call extractor: the extractor is responsible for extracting the required data from the raw data and storing it categorically for future needs. This step enhances efficiency and reduces redundant work. 2. call retriever: the retriever is responsible for recalling and assembling the metadata needed to complete the task from the stored data. This ensures that all necessary information is prepared and ready. 3. call painter (not necessary): the painter is responsible for handling complex data analysis and chart generation. It uses the data extracted and assembled in the previous steps to produce detailed analytical reports and intuitive charts, helping users better understand and interpret the data. Task list TL is a comprehensive series of tasks for each agent need to complete, with each task being derived from the robust DB-GPT resource library which encompasses a wide array of tools, including data extraction, data analysis, and more, designed to support a variety of complex operations and workflows. 

Self-reflection process~\citep{li2024personalllmagentsinsights}. The planner primarily addresses two main aspects~\citep{chen2024autoagentsframeworkautomaticagent}: A. verifying the rationality of the agent team workflow. The following aspects are primarily checked: 1. compliance, ensuring that the agent team adheres to the specifications defined in the constraints. 2. scenario compatibility, verifying that the agent team is compatible with the scenario information defined in the description. 3. system-individual coupling, assessing whether each agent can adapt to the entire team and determining if there is a need to add or remove any agents. B. verifying the rationality of each agent's task list. The following aspects are primarily checked: 1. task interdependency: ensuring that tasks are combined in a logical manner to accomplish specific tasks. 2. input and output parameter logic: verifying that the input and output parameters of each task are logically consistent and that the task can be executed successfully to complete its assigned responsibilities. 

Memory mechanism~\citep{zhang2024surveymemorymechanismlarge}. ROMAS leverages a memory mechanism based on DB-GPT to facilitate communication and feedback between agents. However, due to the varying recency, importance, and relevance of different messages, as well as the limited prompt length processing capability of LLM, it is impractical to cover all messages in a single pass. To address this, ROMAS employs an effective memory categorization strategy, including sensory memory, short-term memory, long-term memory, and hybrid memory, ensuring efficient management and utilization of diverse types of information. Details of the memory mechanism implementation are in the appendix\ref{memory_mechanism}.

\subsection{Execution Phase}\label{sec:execute}
\begin{figure}[ht] 
\centering
\includegraphics[width=\linewidth]{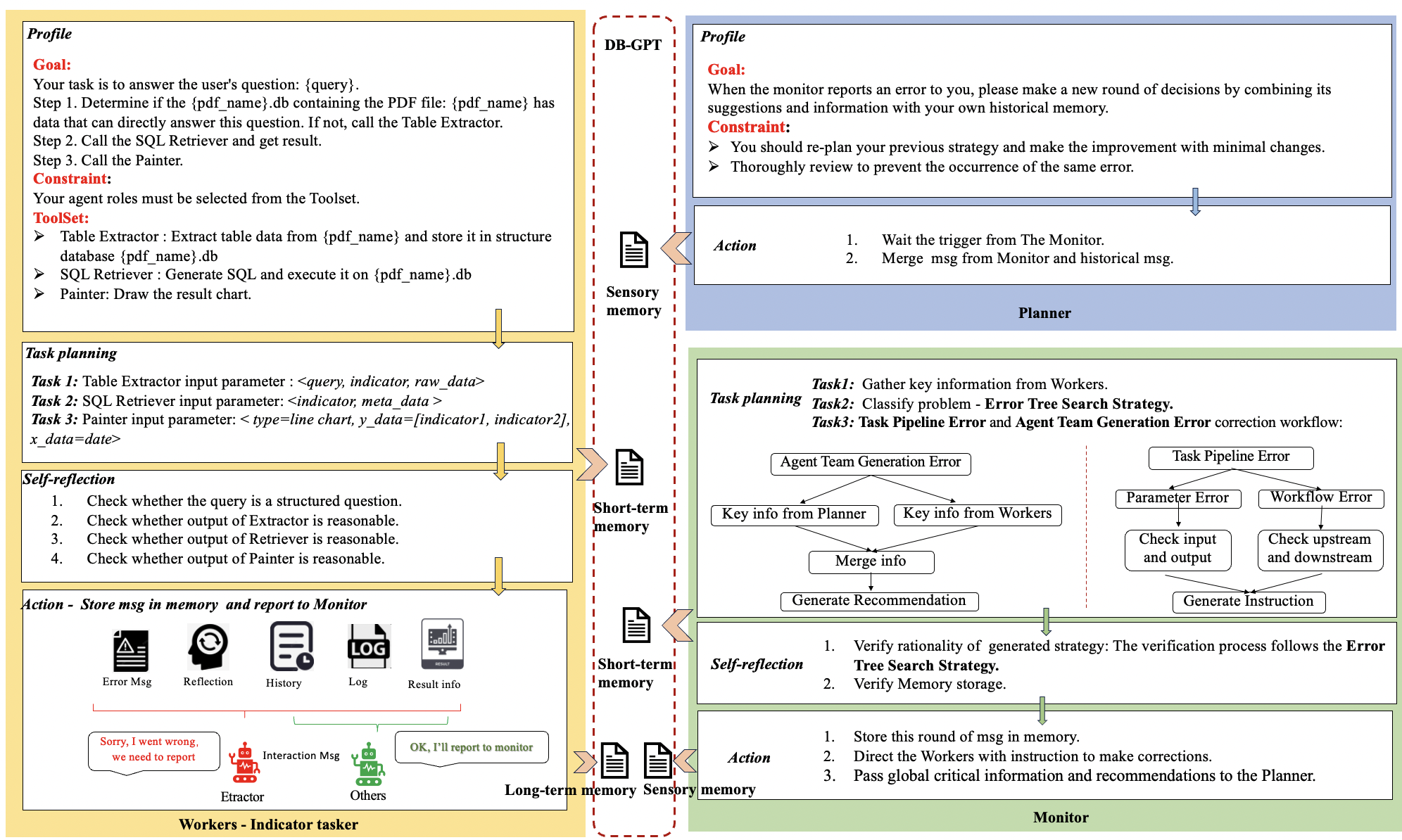} 
\caption{Execution phase of ROMAS, workers execute tasks based on the strategies formulated by the planner, monitor classifies errors and decides to either fix them directly or report to the planner for re-planning.} 
\label{fig:execute_stage} 
\end{figure}
In this phase as illustrated in figure \ref{fig:execute_stage}, When encountering errors, workers firstly attempt to self-correct using the self-reflection mechanism. If the error persists after retries, workers must report the global state to the monitor and await corrective instructions. Upon receiving the urgent information from workers, the monitor firstly classifies the errors. Errors can be classified into two types depending on nature: task list pipeline errors or agent team generation errors. For task list pipeline errors, which typically indicate a problem at a specific process node and regarded as relatively simple orchestration issues, the monitor can directly identify the fault and propose appropriate corrective instructions to the workers. For agent team generation errors, which typically indicates that there exists fundamental system issues in the planning process conducted by the planner. In this case, the monitor needs to conduct a comprehensive analysis of the overall information, formulate a set of improvement recommendations, and trigger the planner to restart the planning process in hopes of finding a better solution.

Error alert. When workers encounter a problem, they report the global state to the monitor. Due to the prompt length window limitation~\citep{vaswani2023attentionneed}, the monitor only processes key information. Therefore, we require that workers encountering errors report detailed information related to both the individual and the system, including logs, error messages, history records, and the self-reflection process. In contrast, workers that are operating normally only need to report their individual result data and system-related context information, such as operation results, history records, and logs. When a worker fails, it will broadcast an error message to all workers, all workers will then pause their tasks and report their current status to the monitor.

Error tree search strategy~\citep{6145622}. Based on empirical data, we have hierarchically organized the potential errors in the MAS into a tree structure, as shown in the appendix\ref{error_tree}. We classify errors into two major categories: pipeline errors and logical errors. Pipeline errors typically involve anomalies in the processing flow, while logical errors are related to defects in algorithm design. When the monitor collects error information, it organizes key information and performs similarity searches using this predefined error tree to pinpoint specific issues, starting from the root node and proceeding layer by layer until it identifies the specific leaf node.  

Instruction and recommendation generation. After identifying the error type, the monitor plans the next steps based on the collected information from both the planner and the workers. (1). Agent team generation errors: The monitor retrieves the planner's planning and reflection record from short-term memory in this current round. It also consolidates alert information provided by the workers, with a particular focus on error-related data. Based on this information, the monitor extracts key global insights and formulates recommendations, which are then transferred to the planner for replanning. (2). Task pipeline error, the monitor will assess whether the input and output results of each task and their upstream and downstream relationships meet the objectives according to the current strategy of the planner. It will then issue adjustment instructions to correct any discrepancies without triggering the planner to perform re-planning.

\subsection{Re-planning Phase}
\begin{figure}[ht] 
\centering
\includegraphics[width=\linewidth]{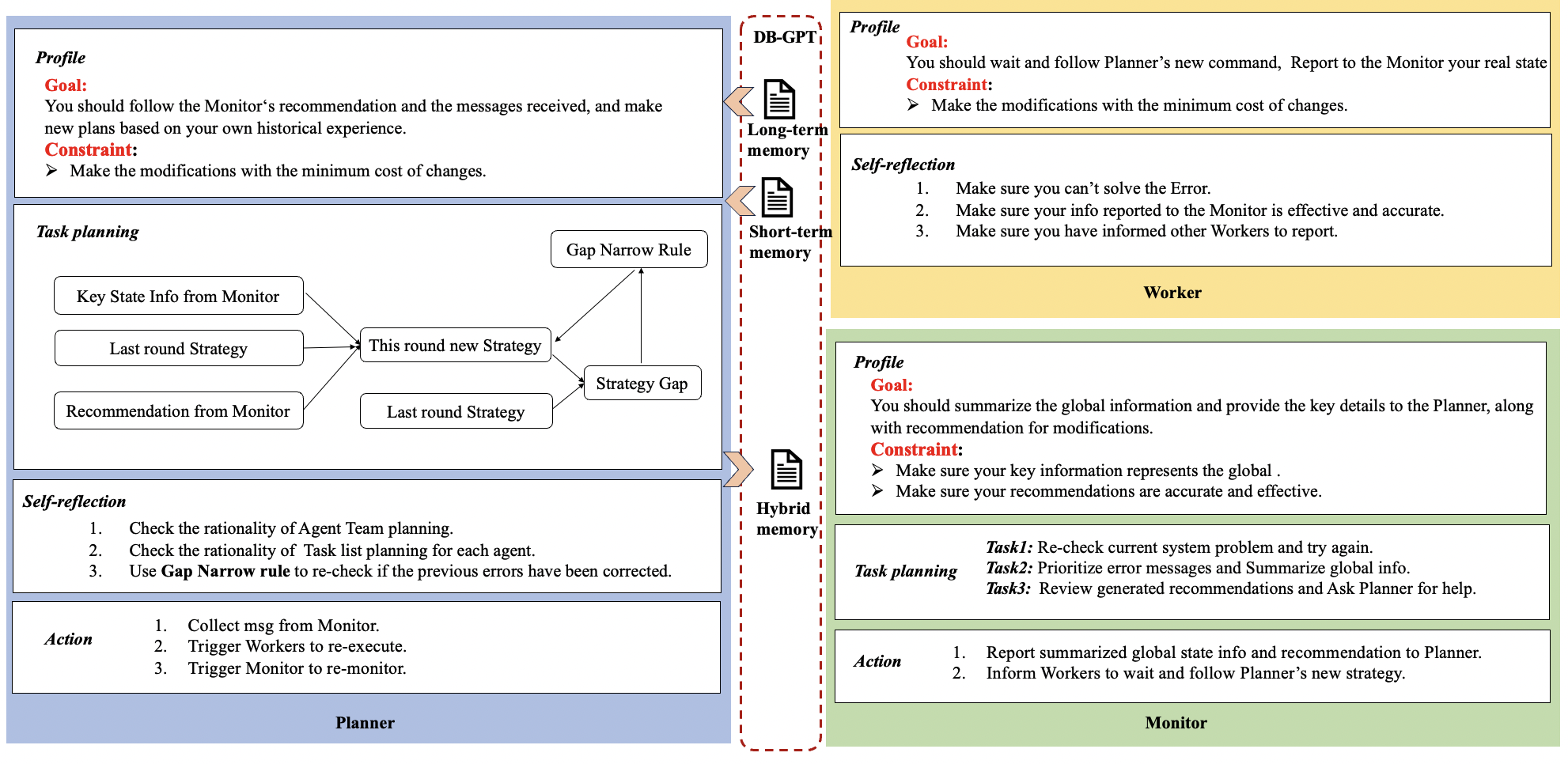} 
\caption{Re-plannig phase of ROMAS, planner receives global critical information and modification recommendation from the monitor to generate a new strategy for the current round.} 
\label{fig:replanning_stage} 
\end{figure}
In this phase as illustrated in figure\ref{fig:replanning_stage}, the main responsibility is on the planner. The planner receives global critical information and modification recommendation from the monitor, and combines them with its own historical strategies and experience information from the previous round to generate a new strategy for the current round. To ensure the effectiveness of the new strategy and avoid resource wastage during the execution stage when validating its effectiveness, we introduce a strategy called gap narrow. The gap narrow strategy aims to correct the errors from the previous round at the minimum cost and align the inconsistencies between the current round's strategy and the previous round's strategy through minimal modifications. This strategy not only enhances the system's efficiency but also ensures the consistency and continuity of the strategies, thereby optimizing the overall performance.  The algorithm at this stage is shown in the algorithm\ref{replan}.  

Gap narrow rule~\citep{li-etal-2024-optimizing-language}. After generating the new strategy, we first use LLM to conduct a detailed comparison between the new and old strategies, analyzing and identifying their specific differences. Next, we establish a set of prior rules aimed at minimizing modifications to the old strategy. Then, we integrate comprehensive data from the monitor to perform a thorough reflection and correction on each difference point. This process not only ensures that each difference point effectively corrects the errors from the old strategy's execution but also aims to achieve the best possible solution with the minimum modification cost~\citep{ZHANG202065}.

\section{Experiments}
\textbf{Datasets}. We conducted our experience based on following 2 dataset, empirical evidence has consistently demonstrated that ROMAS exhibits excellent performance in both general-knowledge and domain-specific scenarios.
\begin{itemize}
    \item FAMMA~\citep{xue2024famma}. We focus on the financial data analysis scenario, a critical application area for generative language models~\citep{xue2023weaverbird,wu2023bloomberggptlargelanguagemodel}, to evaluate the capabilities of ROMAS. FAMMA is an open-source benchmark for financial multilingual multimodal~\citep{multimodal} question answering (QA)~\citep{kapoor2024aiagentsmatter}. To adapt to the task requirements, we processed the original dataset by selecting 100 cases that include both text and table images in the input and have standard options in the output, and converted these table images into tabular format. 
    \item HotpotQA~\citep{yang-etal-2018-hotpotqa}. To demonstrate reasoning capabilities in general scenarios, we selected 100 samples from HotpotQA. This dataset is renowned for its multi-hop reasoning questions and diverse question types, which encourage models to perform cross-document information integration and complex reasoning, thereby comprehensively evaluating and enhancing system performance in complex reasoning tasks.  
\end{itemize}

\textbf{Evaluation metrics}. We select success rate, LLM evaluation~\citep{wang2023largelanguagemodelsfair}, and Human evaluation as metrics. Besides using success rate to evaluate whether the system outputs standard answers, we also adopt CoT to guide the agent in outputting its reasoning process along with standard options. LLM evaluation and Human evaluation are used to assess the accuracy, coherence, completeness, and logic of the descriptions~\citep{wang2023largelanguagemodelsfair}. As shown in the appendix\ref{dimension}, both LLM and human evaluation are scored based on a standard scale of 10 points per dimension, with a total of 100 points. The human score is independently evaluated by multiple financial analysis experts according to a unified standard, with reasons for the scores recorded and the final score averaged. By combining these three metrics, we can comprehensively assess the performance of ROMAS and validate its effectiveness.  

\textbf{Setup}. We developed ROMAS based on GPT-4~\citep{openai2024gpt4technicalreport}, configuring the temperature to 0 to ensure the consistency of the model's outputs. For each agent, the maximum retry count for self-reflection and self-planning was set to 2, mitigating the risk of failure in a single invocation of the LLM. The maximum retry count for the re-planning process was set to 3 to prevent exceeding context threshold.

\textbf{Analysis I: Comparison with other LLM and MAS}. As shown in table \ref{tab:famma_res}, we categorize the comparative objects into three groups: LLMs, single-agent with task planning capabilities, and the current advanced MAS. In the LLMs group, we selected Qwen2-72B~\citep{yang2024qwen2technicalreport}, Llama2-70B~\citep{touvron2023llama2openfoundation}, and GPT-4, combining with single prompt technique as the baselines for experiments. The result indicates that GPT-4 performs the best, likely due to its extensive training with large-scale data and network parameters. In the single-agent group, we selected traditional decision-making models with task planning capabilities, including CoT, ToT, and ReAct. The result shows that ReAct significantly improves the baseline accuracy of LLMs. This improvement is likely because the thought-act-observation~\citep{yao2023reactsynergizingreasoningacting} process in ReAct enables a more reflective and reasonable task planning process. In the MAS group, we compare the pioneering generative agent, which introduces reflective thinking mechanisms in the MAS domain, and AutoAgent, which also features role-based supervision. The result demonstrates that ROMAS outperforms the others. The advantage over the generative agent could be attributed to the introduction of the monitor mechanism, which offers error correction opportunities. ROMAS surpasses AutoAgent possibly because its task planning error correction occurs during the execution phase when real problems arise, rather than during the drafting phase as a prediction. Furthermore, ROMAS's on-the-spot error correction mechanism for pipeline errors significantly enhanced the efficiency of error correction in the MAS. Simultaneously, experimental results show that performance on the HotpotQA dataset surpasses that on FAMMA, particularly within the MAS grouping. This could be attributed to the LLM's proficiency in handling general knowledge, as well as ROMAS can flexibly adjust strategies in the dynamical situation based on self-monitoring and self-planning process, which helps in handling the uncertainties and changes that may arise during multi-hop reasoning.  

\begin{table}[ht]
\centering
\setlength{\tabcolsep}{5pt}
\small
\begin{tabular}{lcccc}
\toprule 
Grouping & Model & $SR$ & $LLMEval$ & $HumanEval$ \\  
 &  & F / H(\%) & F / H(0-100) & F / H(0-100)\\
\midrule 
\multirow{3}{*}{LLM baseline}& LLAMA2-70B & 29.13 / 31.53 & 42.14 / 44.12 & 33.12 / 35.97 \\
& QWEN2-72B & 35.65 / 37.83 & 46.97 / 45.02 & 41.51 / 42.56\\
& GPT-4 & 42.85 / 45.36 & 48.72 / 47.20 & 47.19 / 48.85\\
\cmidrule(lr){1-5}
\multirow{3}{*}{LLM w/ task planning} & CoT (GPT-4) & 46.44 / 51.27 & 51.36 / 54.77 & 50.24 / 49.05\\
& ToT (GPT-4) & 51.61 / 56.18 & 56.98 / 61.04 & 54.79 / 52.61\\
& ReAct (GPT-4) & 56.80 / 61.44 & 57.82 / 63.29 & 60.26 / 62.07\\
\cmidrule(lr){1-5}
\multirow{3}{*}{State-of-art MAS} & Generative Agent & 61.20 / 67.08 & 64.31 / 69.37& 62.03 / 66.07\\
 & AutoAgents & 73.45 / 78.99& 70.55 / 73.08 & 68.07 / 71.11\\
 & \textbf{ROMAS} & \textbf{81.68} / \textbf{85.24}& \textbf{78.30} / \textbf{83.64}& \textbf{75.08} / \textbf{77.98}\\
\hline
\end{tabular}
\vspace{0.3cm}  
\caption{Performance of ROMAS on FAMMA benchmark comparing with other LLM and MAS (F = FAMMA, H = HotpotQA).}
\label{tab:famma_res}
\end{table}

\textbf{Analysis II: Ablation study}. As shown in table \ref{tab:ablation_res}, a comparison of ROMAS performance with and without the corresponding components validates the effectiveness of each component. 

On FAMMA. The result indicates that the absence of the monitor mechanism leads to the most significant decline in ROMAS success rate, underscoring its critical role. Secondly, the self-reflection module also has a considerable impact on the system, although its influence is less than that of the monitor mechanism. The reason for this may be that the self-reflection process can only accomplish independent reflection by the agent itself, based solely on predefined conditions and its own data. In contrast, the information from the monitor mechanism is derived from a global perspective, enabling it to improve the performance of individual agents by considering the overall system state, making the correction process more reliable. The memory mechanism also has a substantial impact on the overall performance of ROMAS, as memory serves as a crucial basis for agents to process information in each round. Without classified storage and prioritization of memory, the task planning capability of agents would significantly decrease. The gap narrow rule has the least impact on ROMAS, indicating that the success rate in the re-planning process after an initial correction is high, thus requiring minimal alignment operations.

Notably, on the HotpotQA, the self-reflection and memory mechanisms have the most significant impact. This disparity may be attributed to the differing inferential demands of the two datasets. The complex questions in FAMMA require precise monitoring mechanisms and self-reflection to avoid errors, whereas HotpotQA's multi-hop reasoning characteristic necessitates a system with stronger memory capacity to maintain consistency and integrity across multiple steps, as well as self-reflection to correct biases in the reasoning process.

\begin{table}[ht]
\centering
\setlength{\tabcolsep}{5pt}   
\small  
\begin{tabular}{lccc}
\toprule 
Model & $SR$ & $LLMEval$ & $HumanEval$ \\  
& F / H $(\% \Delta vs.romas )$ & F / H (0-100) & F / H (0-100)\\
\midrule 
ROMAS w/o gap narrow rule & 76.69 ($\downarrow$ 4.99) / 79.17 ($\downarrow$ 6.07) &  72.76 / 70.61 & 70.85 / 69.29\\
ROMAS w/o memory mechanism & 68.63 ($\downarrow$ 13.05) / 64.83 ($\downarrow$ 20.41)& 65.63 / 68.37 & 62.50 / 68.45\\ 
ROMAS w/o self-reflection & 62.37 ($\downarrow$ 19.31) / 59.92 (\textbf{$\downarrow$25.32})&  54.73 / 45.18  & 52.13 / 41.20\\ 
ROMAS w/o monitor mechanism & 59.02 (\textbf{$\downarrow$22.66}) / 71.79 ($\downarrow$ 13.45)& 42.31 / 53.20 & 40.02 / 50.18\\  
\textbf{ROMAS} & \textbf{81.68} /  \textbf{85.24} &  \textbf{78.30} / \textbf{83.64} & \textbf{75.08} / \textbf{77.98}\\ 
\hline
\end{tabular}
\label{tab:ablation_res}
\vspace{0.3cm}  
\caption{The ablation study of ROMAS (F = FAMMA, H = HotpotQA).}
\end{table}
 
\textbf{Analysis III: DB-GPT effectiveness demonstration}. As shown in table \ref{tab:DBGPT_compare_res}, we validate the effectiveness of developing the ROMAS system using the DB-GPT framework through comparative experiments. We also implemente the ROMAS system using two leading application frameworks, LangChain and AgentScope, and select code volume, average QA time, and task success rate as the evaluation metrics~\citep{feng2020codebertpretrainedmodelprogramming}. The result indicates that the DB-GPT framework significantly reduces the amount of code required for development. This is primarily due to the robust open-source community of DB-GPT, which has contributed numerous functions and operators that support one-click invocation, greatly simplifying the development process of agent applications. Moreover, DB-GPT’s extensive set of database operation tools and components further enhances the success rate of task execution and execution efficiency.  

\begin{table}[ht]
\centering
\setlength{\tabcolsep}{2pt}   
\small  
\begin{tabular}{lccc}
\toprule 
Application framework & $SR$ & $Code Volume $ & $Average QA Time$\\  
& (\%) & (number of rows) & (second) \\
\midrule 
LangChain & 68.59 & 2500 & 22.08 \\
AgentScope & 74.1 & 1800 & 19.97 \\ 
\textbf{ROMAS} & \textbf{81.68} & \textbf{1500} & \textbf{12.23} \\ 
\hline
\end{tabular}
\label{tab:DBGPT_compare_res}
\vspace{0.3cm}  
\caption{ Comparison of ROMAS implementation using DB-GPT, LangChain, and AgentScope.}
\end{table}

\begin{figure}[ht] 
\centering
\includegraphics[width=\linewidth]{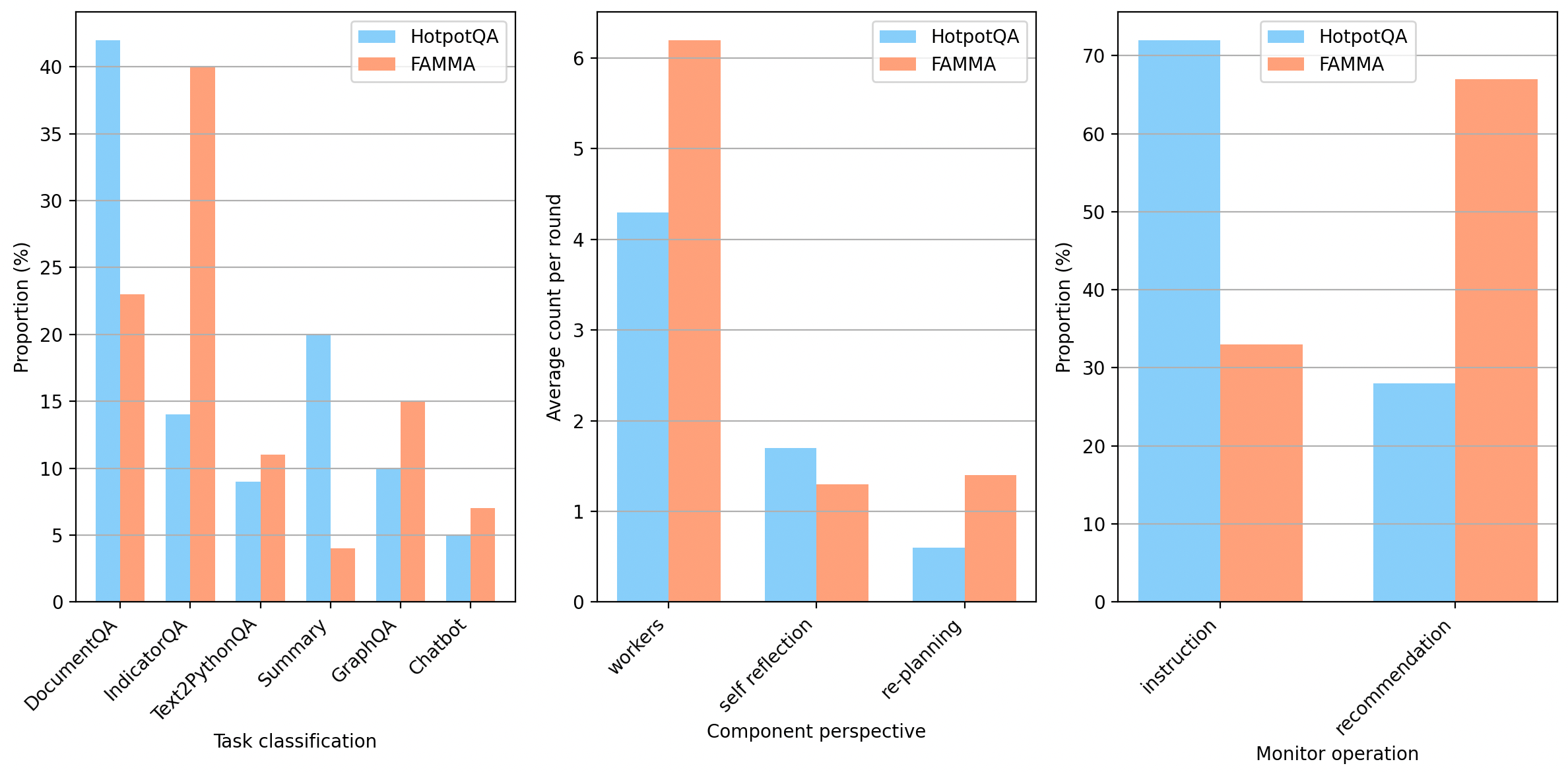} 
\caption{Figure 1 shows the average proportions of different subtask types in the FAMMA and HotpotQA. Figure 2 presents the average number of workers generated, the average self-reflection frequency per agent, and the average replanning frequency of the planner. Figure 3 displays the average proportions of instructions and recommendations given by the monitor.} 
\label{fig:comp_bar} 
\end{figure}

\textbf{Analysis IV: Argument on diversity and functionality}. As shown in figure\ref{fig:comp_bar}, 
the six subtask categories predominantly center on DocumentQA~\citep{documentqa} and IndicatorQA, underscoring their importance in building a qa system and substantial influence on overall system performance. Additionally, for domain-specific dataset FAMMA, the higher task complexity, various data, and complex reasoning necessitate more specialized subtasks, frequent self-reflection and replanning. Consequently, the average frequency of self-reflection and replanning is higher, and the number of generated workers is also greater.

\section{Conclusion}

In this paper, we presented ROMAS, a role-based multi-agent system designed for database monitoring and planning, leveraging DB-GPT for enhanced self-monitoring, self-planning, and collaborative interaction. By addressing the limitations of current multi-agent systems, ROMAS enables efficient and versatile deployment in complex scenarios. Through evaluations on the FAMMA dataset, we demonstrated the system's effectiveness, highlighting its potential to streamline analytical tasks and support future advancements in intelligent multi-agent systems.

\clearpage
\clearpage
\appendix
\appendixpage
\section{Role Introduction}\label{role_intro}
The planner aims to decompose user requests and scenario information into clear, well-defined subtasks, generates the entire specialized agent team ~\citep{chen2024autoagentsframeworkautomaticagent}, and assigns a task list for each agent.

The monitor is responsible for overseeing the entire system, ensuring global smooth operation through continuous interaction with workers and the planner. If an error occurs during execution, the monitor analyzes global information to categorize this error as either task list pipeline error or agent team generation error. Depending on the type of error, monitor decides whether to correct the error itself or to report it to the planner for replanning. 

To align with the typical processes of data acquiring, data cleanning, data processing, and data analysing~\citep{MAHARANA202291} in a data analysis scenario, we have categorized workers into the following roles:

Tasker, as the primary manager for subtasks, aims to complete subtasks by flexibly planning the collaboration between extractors, retrievers, and painters. One system can abstract different types of taskers based on various subtasks, such as indicator tasker, document QA tasker, GraphRAG tasker~\citep{peng2024graphretrievalaugmentedgenerationsurvey}, and summary tasker. 

Extractor, as a data processing assistant, aims to achieve information extraction and index storage from  structured and unstructured raw data. Its functions include PDF document loading, data preprocessing, document block segmentation, document tree construction, table extraction and merging, and the storage of text and table data. 

Retriever, as a data acquiring helper, aims to retrieve the necessary data from the most suitable database to support taskers. Its functions include sql generation, sql execution, python execution,  composite index calculation.

Painter, as a data analysing helper, aims to draw relevant charts based on the provided data, making it easier for users to intuitively understand the analysis results.

\section{Memory Mechanism of ROMAS}\label{memory_mechanism}

Sensory memory~\citep{Wang_2024} is similar to human transient memory and primarily used for recording and capturing real-time sensory information interacted with environment such as one-time and repetitive actions of agents. Some important parts in Sensory memory is transferred to Short-term memory over time. 

Short-term memory~\citep{Wang_2024} stores recent important information with limited capacity and duration. This type of memory temporarily holds information that requires quick access and processing, such as the agent's current self-planning and self-reflection results, agent current situations, context information, temporary strategy. Some important parts in short-term memory is transferred to long-term memory over time. 

Long-term memory~\citep{Wang_2024} stores the knowledge and patterns learned by the agent from past experiences, which are used to guide future decisions and actions. In ROMAS, long-term memory commonly stores agent's vital erroneous information and summaries of historical decisions.

Hybrid memory~\citep{Wang_2024} explicitly combines the advantage of short-term and long-term memories, leveraging immediate data for short-term tasks while utilizing accumulated knowledge for long-term strategy and learning. In ROMAS, hybrid memory is commonly used to generate the current round strategy based on the previous round's strategy and the current round's state, such as in the replanning phase, the planner combines its historical strategy in long-term memory with the global state in monitor's short-term memory to generate this round's new strategy.  

\section{Dimension of LLM and Human Evaluation}\label{dimension}

\begin{figure}[ht]
\centering
\includegraphics[width=0.5\linewidth]{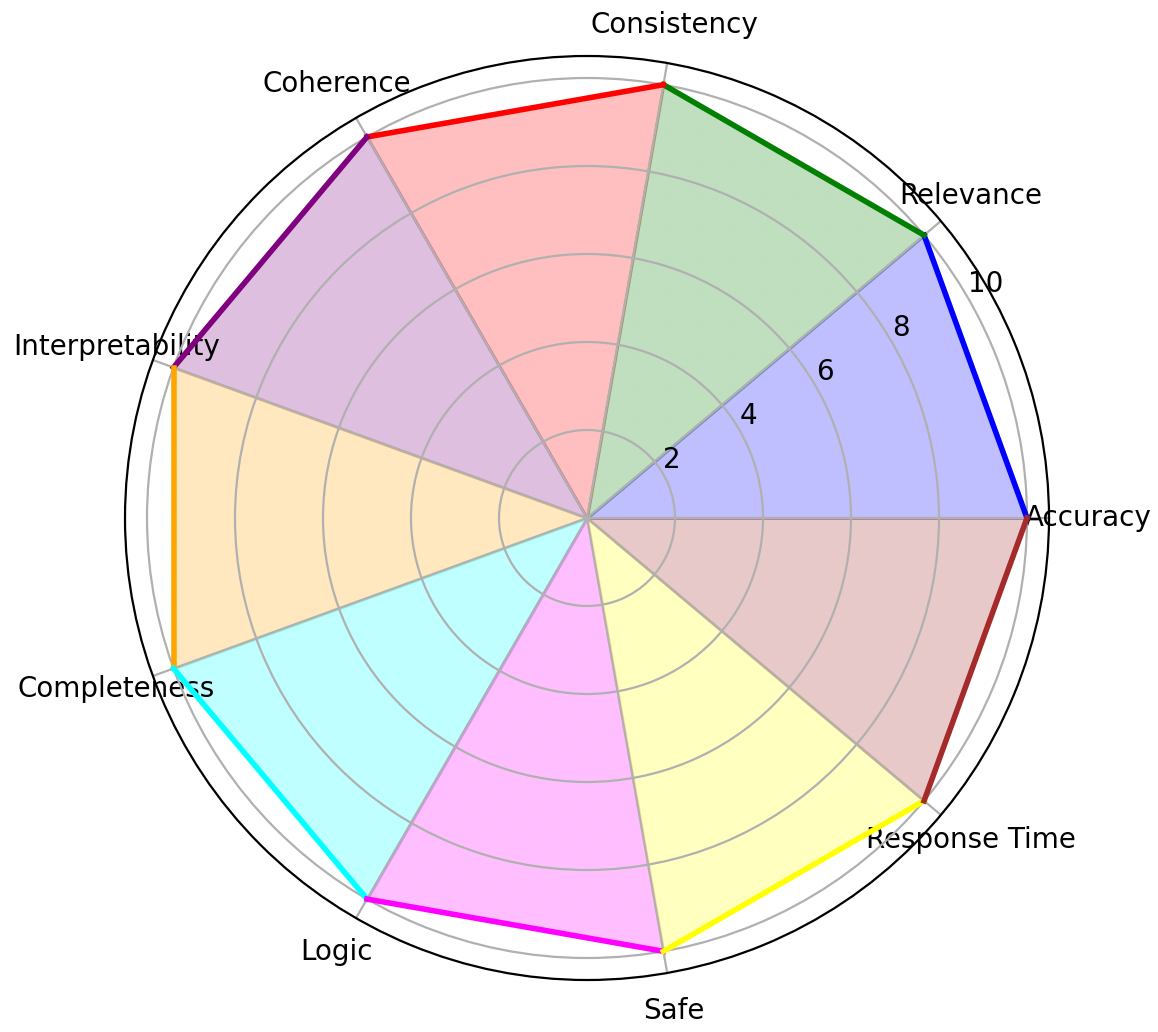} 
\caption{We divided the evaluation criteria into 10 dimensions based on common standards for large models.} 
\end{figure}

\section{Error tree search}\label{error_tree}
\begin{figure}[ht] 
\centering
\includegraphics[width=10cm]{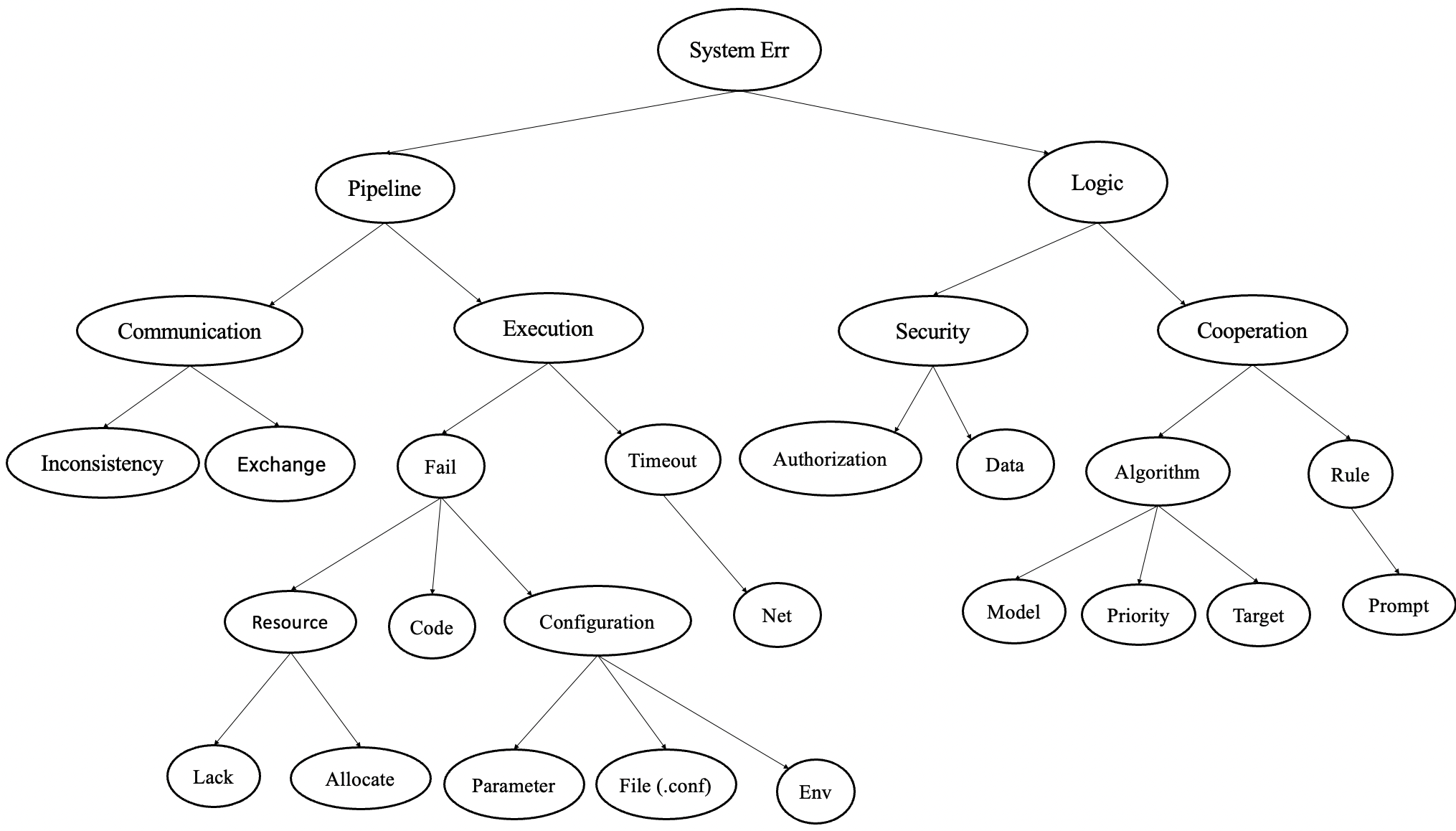}
\caption{Error tree search, we classified common errors in MAS into two categories based on empirical data: pipeline and logic and constructed an error tree as reference for monitor's error classification, enabling precise error cause identification by DFS~\citep{dfs}.}  
\end{figure}

\begin{algorithm}[ht]
\caption{Re-planning phase}
\label{replan}
\begin{algorithmic}[1]
\Require global state from monitor $G$, recommendation from monitor $R$, user query $Q$, old strategy $S_{old}$, scenario prompt $P$
\State generate a new strategy, $S_{new} \leftarrow \text{LLM}(G, R, Q, S_{old}, P)$.
\State compute differences $D$ between $S_{new}$ and $S_{old}$, $D = \{d_1, d_2, \ldots, d_n\} \leftarrow \text{LLM}(S_{new}, S_{old})$.
\State establish prerequisites $R = \{r_1, r_2, \ldots, r_n\}$ aimed at minimizing modifications.

\For{each difference $d_i \in D$}
    \For{each rule $r_j \in R$}
        \If{$d_i$ does not follow $r_j$}
            \State regenerate $d_i$ based on $\{G, R, Q, S_{old}, P\}$
            \State \textbf{break}
        \Else
            \State reflect on whether $d_i$ has improved $S_{old}$ based on $G$ and $R$
            \If{not improved}
                \State regenerate differences $D$ based on $\{G, R, Q, S_{old}, P\}$
                \State \textbf{break}
            \EndIf
        \EndIf
    \EndFor
\EndFor

\State ensure each corrected difference integrates properly to form the optimized strategy $S_{opt}$
\State \Return optimized strategy $S_{opt}$
\end{algorithmic}
\end{algorithm}
\section{Re-planning phase algorithm}

\section{Ongoing and Future Works}
We are actively exploring several extensions to enhance the capability and versatility of our system, particularly in addressing more complex dialogue and analytical tasks. Our key areas of focus include:

\begin{itemize}[leftmargin=*] \item Empowering computational agents.
Users increasingly expect systems not only to perform analyses but also to deliver advanced computational capabilities, such as generating predictive insights ~\citep{jin2023large,xue2024easytpp} and facilitating decision-making based on historical data~\citep{xue_meta_2022,pan2023deep,zhou2023gmp}. Developing these functionalities will significantly enhance the system's utility in real-world applications, such as recommendation systems~\citep{chu2023leveraging,wang2023llmrg} and traffic control~\citep{xue2022hypro}.

\item Adopting advanced model training methodologies.
Beyond pre-training, the integration of advanced techniques such as continual learning, including continual pre-training~\citep{jiang2023anytime,jiang2024interpretable}, and prompt learning~\citep{wang2022learning,xue2023prompttpp}, offers opportunities to improve the system's adaptability and performance. Leveraging these methods will drive advancements in the development of systems for computational use and foster further innovation in the research community.
\end{itemize}
\bibliographystyle{plainnat}
\bibliography{references}
\end{document}